\documentclass[11pt,a4paper]{article}
\PassOptionsToPackage{hyphens}{url}\usepackage{hyperref}    
\usepackage[hyperref]{acl2020}
\usepackage{times}
\usepackage{latexsym}
\usepackage{graphicx}
\usepackage[normalem]{ulem}
\usepackage{enumitem}
\usepackage{listings}
\usepackage{amsmath}
\usepackage{url}
\usepackage{xspace}
\usepackage{booktabs}
\usepackage{textcomp}
\usepackage{colortbl}
\usepackage{makecell}

\aclfinalcopy 
\def\aclpaperid{2020.acl-main.447} 

\newcommand{\textapprox}{\raisebox{0.5ex}{\texttildelow}}

\newcommand{\cord}{\textsc{CORD\nobreakdash-19}\xspace}
\newcommand{\bert}{\textsc{BERT}\xspace}
\newcommand{\scibert}{\textsc{SciBERT}\xspace}

\newcommand{\grobid}{\textsc{Grobid}\xspace}
\newcommand{\gorc}{\textsc{S2ORC}\xspace}
\newcommand{\latex}{\textsc{LaTeX}\xspace}
\newcommand{\gorbert}{\textsc{S2ORC-SciBERT}\xspace}
\newcommand{\semanticscholar}{Semantic Scholar\xspace}

\colorlet{punct}{red!60!black}
\definecolor{background}{HTML}{F9F9F9}
\definecolor{delim}{RGB}{20,105,176}
\definecolor{dartmouthgreen}{rgb}{0.05, 0.5, 0.06}
\colorlet{numb}{black!60!black}

\lstdefinelanguage{json}{
    basicstyle=\normalfont\ttfamily\small,
    numbers=left,
    numberstyle=\scriptsize,
    stepnumber=1,
    numbersep=2pt,
    showstringspaces=false,
    breaklines=true,
    frame=lines,
    backgroundcolor=\color{background},
    literate=
     *{0}{{{\color{numb}0}}}{1}
      {1}{{{\color{numb}1}}}{1}
      {2}{{{\color{numb}2}}}{1}
      {3}{{{\color{numb}3}}}{1}
      {4}{{{\color{numb}4}}}{1}
      {5}{{{\color{numb}5}}}{1}
      {6}{{{\color{numb}6}}}{1}
      {7}{{{\color{numb}7}}}{1}
      {8}{{{\color{numb}8}}}{1}
      {9}{{{\color{numb}9}}}{1}
      {:}{{{\color{punct}{:}}}}{1}
      {,}{{{\color{punct}{,}}}}{1}
      {\{}{{{\color{delim}{\{}}}}{1}
      {\}}{{{\color{delim}{\}}}}}{1}
      {[}{{{\color{delim}{[}}}}{1}
      {]}{{{\color{delim}{]}}}}{1},
}

\setcitestyle{round}

\title{S2ORC: The Semantic Scholar Open Research Corpus}

\author{
    Kyle Lo$^{\dagger}$\Thanks{ denotes equal contribution}
    \quad Lucy Lu Wang$^{\dagger}$\footnotemark[1]
    \quad Mark Neumann$^{\dagger}$
    \quad Rodney Kinney$^{\dagger}$
    \quad Daniel S. Weld$^{\dagger \ddag}$  \vspace{2pt} \\
    $^\dagger$Allen Institute for Artificial Intelligence \vspace{2pt} \\
    $^\ddag$Paul G. Allen School of Computer Science \& Engineering, University of Washington \vspace{2pt} \\
    \texttt{\{kylel, lucyw\}@allenai.org}
}

\date{}

\begin{document}

\maketitle
\begin{abstract}
We introduce \gorc,\footnote{Instructions for access to the data and model are available at \url{https://github.com/allenai/s2orc/}.} a large corpus of 81.1M English-language academic papers spanning many academic disciplines.  The corpus consists of rich metadata, paper abstracts, resolved bibliographic references, as well as structured full text for 8.1M open access papers. Full text is annotated with automatically-detected inline mentions of citations, figures, and tables, each linked to their corresponding paper objects. In \gorc, we aggregate papers from hundreds of academic publishers and digital archives into a unified source, and create the largest publicly-available collection of machine-readable academic text to date. We hope this resource will facilitate research and development of tools and tasks for text mining over academic text. 
\end{abstract}

\section{Introduction}

Academic papers are an increasingly important textual domain for natural language processing (NLP) research.
Aside from capturing valuable knowledge from humankind's collective research efforts, academic papers exhibit many interesting characteristics -- thousands of words organized into sections, objects such as tables, figures and equations, frequent inline references to these objects, footnotes, other papers, and more.  

\begin{figure}[t!]
    \centering
    \includegraphics[width=\columnwidth]{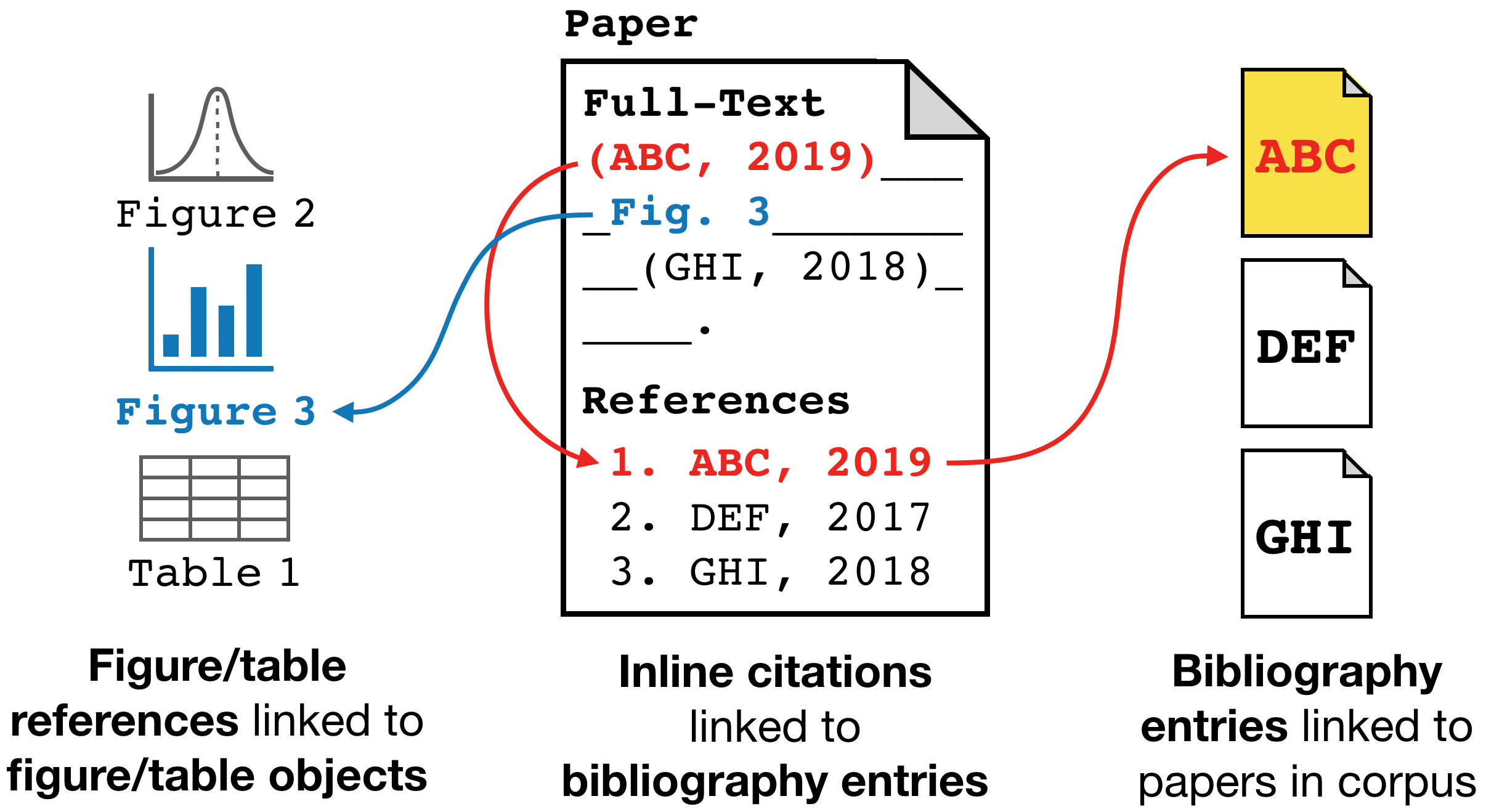}
    \caption{Inline citations and references to figures and tables are annotated in \gorc's structured full text. Citations are linked to bibliography entries, which are linked to other papers in \gorc.  Figure and table references are linked to their captions.}
    \label{fig:gorc_links}
\end{figure}

\begin{table*}[th!]
    \centering
    \scalebox{0.78}{
        \begin{tabular}{p{55mm}p{20mm}p{18mm}p{28mm}p{27mm}p{30mm}}
            \toprule
            Corpus	 &	\makecell[l]{Papers w/ \\ body text}	&	\makecell[l]{Citation \\ contexts} & \makecell[l]{References to \\ tables / figures / \\ equations}	&
            \makecell[l]{Linked to \\ graph} & \makecell[l]{Academic \\ disciplines}	\\
            \midrule
            \textbf{\gorc (PDF-parse)}	& \textbf{8.1M}	&	\textbf{full text}	&	\textbf{yes}	&	\textbf{\gorc (full)} & \textbf{multi}	\\
            \textbf{\gorc (\latex-parse)} &	\textbf{1.5M}	&	\textbf{full text}	&	\textbf{yes}	&	\textbf{\gorc (full)} & \textbf{physics, math, CS}	\\
            PubMed Central (OA)	& 	2.6M	&	full text	& 	yes	&	PubMed  & bio, med	\\
            AAN	\cite{Radev:2009:AAN:1699750.1699759} &	25k  &	full text	&	no	& ACL Anthology	& comp ling	\\
            \arrayrulecolor{black!10}\midrule
            \citet{Saier2019BibliometricEnhancedAA}$^\dagger$ & 1.0M	&	snippets	&	no	& MAG	 & physics, math, CS	\\
            RefSeer	\cite{Huang2015ANP} &	1.0M	&	snippets	&	no	& CiteSeerX	& multi	\\
            \arrayrulecolor{black}\bottomrule
        \end{tabular}
    }
    \caption{
        A comparison of \gorc with other publicly-available academic text corpora. 
        Of the other corpora: PubMed Central (OA) links to PubMed, which contains 30M papers at the time of writing. AAN links to the ACL Anthology (which contained 25k papers at the time of dataset construction, and 54k papers at the time of writing). \citet{Saier2019BibliometricEnhancedAA} is derived from arXiv and links to MAG (which contained 213M papers and other non-paper documents at the time of dataset construction, and 226M nodes at the time of writing). RefSeer links to CiteSeerX (which contained 1M papers at the time of dataset construction, and 6M papers at the time of writing).
        \gorc contains three times more full text papers than PubMed Central (OA), the next largest corpus with bibliometric enhancements, while covering a more diverse set of academic disciplines. 
        Citations in \gorc are linked to the full set of \gorc papers, 81.1M paper nodes derived from \semanticscholar.
        In addition, the \latex subset of \gorc captures additional structure omitted by \citet{Saier2019BibliometricEnhancedAA}, who also parse \latex sources from arXiv. \\ [1mm]
        \footnotesize{$^\dagger$\citet{Saier2020} is an update to this work which now includes full text.  It is released concurrently with this work.}
    }
    \label{tab:other_corpora}
\end{table*}

Different types of resources have been used to support research over academic papers. Citation graphs like AMiner's Open Academic Graph \cite{Tang2008ArnetMinerEA}, the Microsoft Academic Graph (MAG) \cite{Shen2018AWS}, and the Semantic Scholar literature graph \cite{Ammar2018ConstructionOT}, have had widespread application in bibliometrics, science-of-science, information retrieval, and network analysis. Digital archives like arXiv,\footnote{\url{https://arxiv.org}} PubMed Central,\footnote{\url{https://www.ncbi.nlm.nih.gov/pmc}} CiteSeerX \citep{Giles1998CiteSeerAA},\footnote{\url{https://citeseerx.ist.psu.edu}} and the ACL Anthology \cite{bird-etal-2008-acl},\footnote{\url{https://www.aclweb.org/anthology}} are popular resources for deriving large text corpora for summarization and language modeling or, with further annotation, development of datasets for tasks like entity extraction, text classification, parsing, and discourse analysis.  We focus on bibliometrically-enhanced derivations of these corpora, such as the ACL Anthology Network (AAN) \cite{Radev:2009:AAN:1699750.1699759}\footnote{\url{http://aan.how/}} derived from the ACL Anthology, RefSeer \cite{Huang2015ANP} derived from CiteSeerX, and \citet{Saier2019BibliometricEnhancedAA} derived from arXiv, which combine useful aspects of citation graphs and raw text corpora.  These resources provide citation mentions linked to paper identifiers in their corresponding digital archives, such as the ACL Anthology and CiteSeerX, or to nodes in citation graphs such as MAG, enabling new forms of cross-paper discourse analysis (e.g., studying \textit{how} or \textit{why} papers are related).

Yet, existing corpora are not without their limitations. Some cover a small number of papers (e.g. AAN), are domain-specific (e.g. AAN, PubMed Central, \citet{Saier2019BibliometricEnhancedAA}), or may not provide usable full text (e.g. \citet{Saier2019BibliometricEnhancedAA} and RefSeer). 
To address these issues, we introduce \gorc,\footnote{pronounced ``stork''} the Semantic Scholar\footnote{The papers included in \gorc are a curated subset of the papers in the \semanticscholar literature graph \cite{Ammar2018ConstructionOT} that focuses only on English-language papers with abstracts or full text available. See \S\ref{sec:filtering_paper_clusters} for details on filtering through Semantic Scholar papers.} Open Research Corpus, a large publicly-available collection of 81.1M academic papers covering dozens of academic disciplines.  Each paper is associated with metadata and abstracts aggregated from hundreds of trusted sources such as academic publishers and literature archives like PubMed and arXiv. 

Notably, we release structured, machine-readable full text extracted from PDFs for 8.1M papers which we've identified as having open access status. \gorc full text preserves meaningful structure, e.g., paragraph breaks, section headers, inline citation mentions, references to tables and figures, and resolved citation links to other papers. Additionally, we provide 1.5M full text \latex parses from which we have extracted, in addition to citations and references, the source text of tables and mathematical formulas. As shown in Table \ref{tab:other_corpora}, \gorc provides substantially more structured full text papers and covers a more diverse set of academic disciplines than other resources.

In this paper, we describe the construction of \gorc (\S\ref{sec:graph_construction}). We provide summary statistics of the corpus (\S\ref{sec:gorc_dataset}) and evaluate the data quality (\S\ref{sec:evaluation}). We then evaluate a \bert model pretrained on \gorc (\S\ref{sec:gorbert}), and discuss potential applications to a variety of NLP and analysis tasks over academic text (\S\ref{sec:discussion}). Finally, we compare \gorc with other publicly-available academic text corpora (\S\ref{sec:related_work}).

\section{Constructing the corpus}
\label{sec:graph_construction}

\gorc is constructed using data from the \semanticscholar literature corpus \citep{Ammar2018ConstructionOT}. 
Papers in \semanticscholar are derived from numerous sources: obtained directly from publishers, 
from resources such as MAG, 
from various archives such as arXiv or PubMed, or crawled from the open Internet.
\semanticscholar clusters these papers based on title similarity and DOI overlap, resulting in an initial set of approximately 200M paper clusters.

To construct \gorc, we must overcome challenges in (i) paper metadata aggregation, (ii) identifying open access publications, and (iii) clustering papers, in addition to identifying, extracting, and cleaning the full text and bibliometric annotations associated with each paper. The pipeline for creating \gorc is: 

\begin{enumerate}[noitemsep,label={\arabic*)},topsep=\parskip]
    \item Process PDFs and \latex sources to derive metadata, clean full text, inline citations and references, and bibliography entries,
    \item Select the best metadata and full text parses for each paper cluster, 
    \item Filter paper clusters with insufficient metadata or content, and
    \item Resolve bibliography links between paper clusters in the corpus. 
\end{enumerate}

\noindent Details for these steps are provided below.  See Appendix \S\ref{appendix:terminology} for definitions of terminology. The output of this pipeline is visualized in Figure \ref{fig:gorc_links}.

\subsection{Processing PDFs}
\label{sec:pdf_processing}

We process PDFs from the \semanticscholar corpus using \textsc{ScienceParse} v3.0.0\footnote{\url{https://github.com/allenai/science-parse}} and \grobid v0.5.5\footnote{\url{https://github.com/kermitt2/grobid}} \cite{Lopez2009GROBIDCA}. Our processing pipeline is described below.

\paragraph{Selecting PDFs} We remove PDFs which are less likely to be academic papers. \textsc{ScienceParse} and \grobid are not optimized for processing non-paper academic documents such as dissertations, reports, slides, etc., and this filtering step is necessary to increase output data quality. See Appendix \S\ref{appendix:pdf_filters} for filter details. There are around 31.3M PDFs associated with approximately 200M initial paper clusters, and 30.5M PDFs are selected for processing based on these filtering criteria.

\paragraph{Extracting structured data from PDFs}
We use \textsc{ScienceParse} to extract title and authors from each PDF.\footnote{Our evaluations suggest \textsc{ScienceParse} outperforms \grobid for title and author extraction.} We then use \grobid to process each PDF. From the XML output of \grobid, we extract (i) metadata such as title, authors, and abstract, (ii) paragraphs from the body text organized under section headings, (iii) figure and table captions, (iv) equations, table content, headers, and footers, which we remove from the body text, (v) inline citations in the abstract and body text, (vi) parsed bibliography entries with title, authors, year, and venue identified, and (vi) links between inline citation mentions and their corresponding bibliography entries.  

\paragraph{Postprocessing \grobid output}

We postprocess \grobid output using regular expressions to classify the parenthetical citation style of a paper as \textsc{bracket} (e.g. \textsc{[2]}), \textsc{name-year} (e.g. \textsc{ABC, 2019}), or \textsc{other} (superscripts and other mixed styles). We focus on addressing two types of common errors in \grobid's inline citation extractions: (i) false positives resulting from superscripts or equation references being recognized as inline citations in papers with \textsc{bracket}-style citations, and (ii) false negatives resulting from an inability to expand bracket citation ranges (e.g. ``[3]-[5]'' should be expanded to ``[3], [4], [5]'' before linking). False positives are detected using regular expressions and removed from \grobid output. Bracket citation ranges are manually expanded and linked to their corresponding bibliography entries. The resulting parses are expressed in JSON format.\footnote{The \gorc data format is described at \url{https://github.com/allenai/s2orc}}

\subsection{Processing \latex source}
\label{sec:processing_latex_source}

\latex document source is available for a majority of arXiv submissions, and where available, are used to construct a full text parse. We retrieve body text, section headers, figure/table captions, table representations, equations, and inline citations and references directly from \latex source. Inspired by \citet{Saier2019BibliometricEnhancedAA}, we first convert \latex source into XML documents and then extract structured information from the XML.

Due to direct access to source, the accuracy of citation span, reference, caption, section header, and equation detection is near-perfect. We process 1.5M papers from \latex source derived from arXiv, all of which are included as part of \gorc. Surprisingly, due to the diversity of ways in which authors define metadata in \latex, the quality of metadata extracted from \latex documents is worse than those extracted from PDF. Therefore, we do not use \latex-derived metadata for paper clustering or metadata selection.

\subsection{Selecting canonical metadata}
\label{sec:canonical_representation}

Canonical values for title, authors and other metadata fields are selected from among the papers in a cluster. 
First, if a cluster contains multiple PDFs, we select one to be canonical. This can occur, for example, in a cluster containing an arXiv preprint and its eventual camera-ready version. We preferentially select PDFs from open access sources and break ties by prioritizing PDFs for which there exist richer publisher-provided metadata (e.g. abstract, year, venue, DOI). 
If the selected PDF is associated with publisher-provided metadata, we select those publisher-provided metadata fields to be canonical. 

In cases where publisher-provided metadata is incomplete, we use majority voting to select canonical metadata values.  We break ties by minimizing the total number of sources from which we select metadata (e.g., if IEEE provides title, authors and abstract, DBLP provides title and authors, and arXiv provides title and abstract, we prioritize selecting IEEE over the union of DBLP and arXiv).
\gorc metadata fields include title, author, year, venue, journal, abstract, and identifiers (DOI, PubMed, PubMed Central (PMC), arXiv, and ACL Anthology).

In cases where the title and authors are not provided by any publishers, we derive the values for these fields from the parsed PDF, prioritizing \textsc{ScienceParse} over \grobid.
We further comment on paper clustering as it pertains to metadata selection in Appendix \S\ref{appendix:clustering_logic}.

\subsection{Assembling the corpus}
\label{sec:assembling_the_corpus}

We construct the final corpus by assembling clustered paper metadata with \grobid and \latex parse objects. We associate the \grobid parse with the \gorc paper object if a valid \grobid parse is produced from the PDF, and the PDF is open access. Open access status is assigned if a paper is derived from arXiv, ACL Anthology, PubMed Central (OA), and/or associated with an open-access DOI in the Unpaywall database.\footnote{Unpaywall 2019-04-19 data dump} If the PDF is not open access, we only include the bibliography from the \grobid parse in \gorc. If arXiv \latex source is available for the paper cluster, we also associate the \latex parse with the \gorc paper object.

\subsection{Filtering paper clusters}
\label{sec:filtering_paper_clusters}

We further filter paper clusters to remove papers with (i) no title, (ii) no authors, (iii) fewer than 100 characters of abstract and body text, and (iv) where English is not the primary language. The first three filters remove papers that provide little value for bibliometric-based or text-based analyses.
The English language filter\footnote{We use the \texttt{cld2} tool for language detection with a threshold of 0.9 over the English language score.} reduces \grobid parsing errors. All filters are applied in series.

Subsequently, 95.5M paper clusters are filtered out based on the aforementioned criteria and removed from the corpus. The distribution of filtered papers is given in Table \ref{tab:filters}. We note that a large number of paper clusters are filtered out; 80.0M of these filtered clusters have no associated publisher-provided abstract or associated PDF and do not provide significant value to our dataset in their current state. Although these papers that lack text may be useful as cite-able nodes in \gorc, they are generally of lower quality and are filtered out of the corpus to improve corpus quality.

\begin{table}[ht]
    \centering
    \begin{tabular}{lc}
        \toprule
        Filter & Number of papers \\
        \midrule
        No title & 20k \\
        No authors & 0.3M \\
        $<$ 100 chars of text & 80.0M \\
        Not English & 15.2M \\
        \bottomrule
    \end{tabular}
    \caption{Post-processing data quality filters for papers}
    \label{tab:filters}
\end{table}

\subsection{Linking bibliographies to papers}
\label{sec:linking_bibliographies_to_papers}

Each bibliography entry in both \grobid and \latex parses are linked to the most similar papers in the corpus. For linking, we score each bibliography entry and paper cluster pair using a similarity score computed between their titles. Each title is first normalized (i.e. white spaces stripped, lower-cased, special characters removed) and represented by its character 3-grams. The similarity score $S_{title}$ is computed as the harmonic mean between a Jaccard index and a containment metric:
\begin{equation}
    S_{title} = \frac{2 \times J \times C}{J + C}
\end{equation}

\noindent where the Jaccard index $J$ and containment metric $C$ are computed from the $n$-grams of the two titles $N_1$ and $N_2$ as:
\begin{equation*}
    J = \displaystyle\frac{|N_1 \cap N_2|}{|N_1 \cup N_2|}
\end{equation*}
\begin{equation*}
    C = \displaystyle\frac{|N_1 \cap N_2|}{\min{(|N_1|,|N_2|)}}
\end{equation*}

For each bibliography entry, the bibliography-paper pair with the highest similarity score above 0.8 is output as the correct link. Otherwise, the bibliography entry remains unlinked. We perform an evaluation of linking performance in \S\ref{sec:evaluation}.

\section{The \gorc dataset}
\label{sec:gorc_dataset}

The resulting corpus consists of 81.1M papers.  Our publisher-provided abstract coverage is 90.4\%, or 73.4M papers.  Our PDF coverage is 35.6\%, or 28.9M papers.  These PDFs are processed using the pipeline discussed in \S\ref{sec:pdf_processing}.  The vast majority of these PDFs are successfully processed using \grobid, and we extract bibliography entries for 27.6M of the 28.9M PDFs.  We identify 8.1M of the 28.9M PDFs as open access (\S\ref{sec:assembling_the_corpus}), and we provide full text for all papers in this open access subset. For the 1.5M papers for which \latex source is available through arXiv, we further obtain and provide \latex parses (\S\ref{sec:processing_latex_source}). Using these extracted bibliographies, we resolve a total 380.5M citation links between papers  (\S\ref{sec:linking_bibliographies_to_papers}), 156.5M of which can be tied back to their inline citation mentions in the full text. See Table \ref{tab:stats_provenance} for more provenance statistics.

\begin{table}[t!]
    \centering
    \begin{tabular}{ll}
        \toprule
        Total papers & 81.1M \\
        \midrule
        Papers w/ PDF & 28.9M (35.6\%) \\
        Papers w/ bibliographies & 27.6M (34.1\%) \\
        Papers w/ \grobid full text & 8.1M (10.0\%) \\
        Papers w/ LaTeX full text & 1.5M (1.8\%) \\
        Papers w/ publisher abstract & 73.4M (90.4\%) \\
        Papers w/ DOIs & 52.2M (64.3\%) \\
        Papers w/ Pubmed IDs & 21.5M (26.5\%) \\
        Papers w/ PMC IDs & 4.7M (5.8\%) \\
        Papers w/ ArXiv IDs & 1.7M (2.0\%)  \\
        Papers w/ ACL IDs & 42k (0.1\%) \\
        \bottomrule
    \end{tabular}
    \caption{Statistics on paper provenance. We note that categories are not mutually exclusive and do not sum to 100\%. All papers in \gorc have either a publisher-provided abstract or an associated PDF from which we derive full text and/or bibliography entries, or both.}
    \label{tab:stats_provenance}
\end{table}

\begin{table}[t!]
    \centering
    \scalebox{0.96}{
        \begin{tabular}{p{40mm}ll}
            \toprule
            Statistic & \grobid & \latex \\
            \midrule
            Paragraphs (abstract) & 1.1 & - \\
            Paragraphs (body) & 9.9 & 93.3* \\
            \midrule
            Inline cite spans (abstract) & 0.7 & - \\
            Inline cite spans (body) & 45.2 & 46.8 \\
            \midrule
            Bibliography entries & 27.6 & 21.9 \\
            Linked bib. entries & 19.3 & 6.8$^\dagger$ \\
            \bottomrule
        \end{tabular}
    }
    \caption{Extraction and linking statistics over PDF and \latex parses. Reported values are averaged over all open access papers, which consist of 8.1M \grobid-parsed PDFs and 1.5M parsed \latex sources. \\ [1mm]
    \footnotesize{
        *\latex preserves line breaks rather than paragraph breaks. \\ [1mm] 
        \small{$^\dagger$}The lower number of linked bibliography entries in \latex parses is due to large numbers of papers (mostly in the field of physics) for which the bibliography entries are formatted without paper titles. Our linking algorithm strongly depends on titles and fails to link these entries.}}
    \label{tab:stats_grobid_latex}
\end{table}

\begin{figure}[tbh!]
    \centering
    \includegraphics[width=\columnwidth]{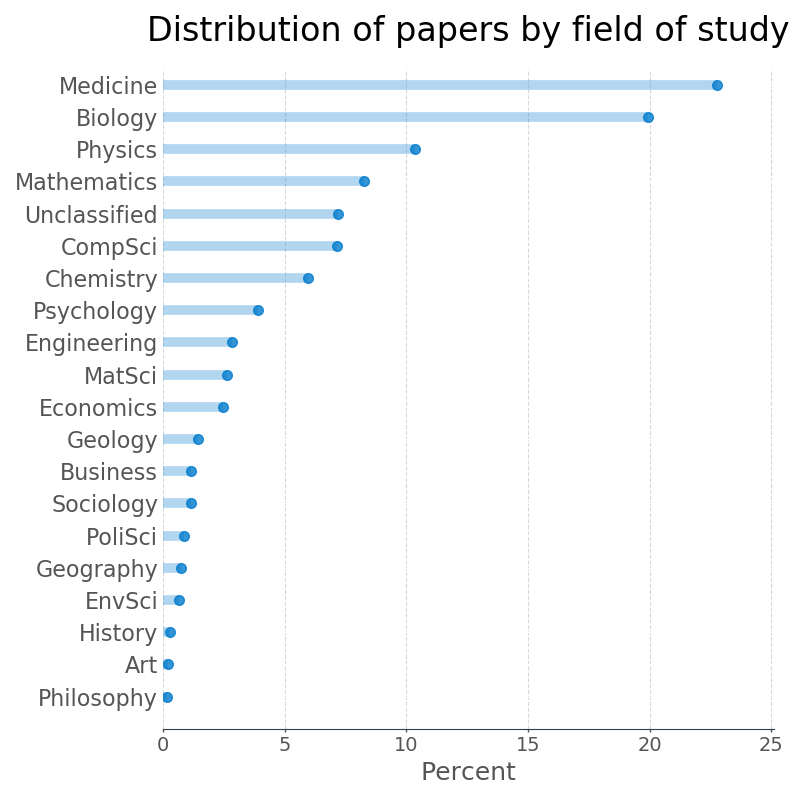}
    \caption{Distribution of papers by Microsoft Academic field of study.}
    \label{fig:fos_stats}
\end{figure}

We provide statistics for the \grobid and \latex full text parses and bibliography linking in Table \ref{tab:stats_grobid_latex}. On average, \latex parses contain many more ``paragraphs'' of body text, because \latex source files preserve line breaks rather than paragraph breaks. We speculate that differences in bibliography entry and linking counts between the \grobid and \latex parses are due to a combination of: (i) challenges in \latex bibliography expansion and parsing, and (ii) differences in bibliography formatting in some math and physics venues (where bibliography entries do not include paper titles, which we depend on for bibliography linking).

The distribution of academic disciplines in \gorc is given in Figure \ref{fig:fos_stats} using Microsoft Academic fields of study. Not all papers in \gorc can be found in Microsoft Academic -- those not found are denoted as \emph{Unclassified}. Approximately 677k papers have more than one primary Microsoft Academic field of study; Figure \ref{fig:fos_stats} represents only the top field of study for each paper.

\begin{table*}[t!]
    \centering
    \scalebox{0.81}{
        \begin{tabular}{p{23mm}p{25mm}p{45mm}p{23mm}p{22mm}p{25mm}}
            \toprule
            Domain & Dataset	 &	Reference	&	Task & \textsc{Sci}BERT	& \gorbert \\
            \midrule
             & BC5CDR & \citet{li2016biocreative} & NER & 90.01 & 90.41 $\pm$ 0.06 \\
             & JNLPBA & \citet{collier-kim-2004-introduction} & NER & 77.28 & 77.70 $\pm$ 0.25 \\
             & NCBI-disease & \citet{dougan2014ncbi} & NER & 88.57 & 88.70 $\pm$ 0.52 \\
            \emph{Biomed} & EBM-NLP & \citet{nye-etal-2018-corpus} & PICO & 72.28 & 72.35 $\pm$ 0.95 \\
             & GENIA & \citet{kim2003genia} & DEP (LAS) & 90.43 & 90.80 $\pm$ 0.19 \\
             & GENIA & \citet{kim2003genia} & DEP (UAS) & 91.99 & 92.31 $\pm$ 0.18 \\
             & ChemProt & \citet{Krallinger2017OverviewOT} & REL & 83.64 & 84.59 $\pm$ 0.93 \\
            \midrule
             & SciERC & \citet{luan-etal-2018-multi} & NER & 67.57 & 68.93 $\pm$ 0.19 \\
            \emph{CS} & SciERC & \citet{luan-etal-2018-multi} & REL & 79.97 & 81.77 $\pm$ 1.64 \\
             & ACL-ARC & \citet{Jurgens2018MeasuringTE} & CLS & 70.98 & 68.45 $\pm$ 2.47 \\
            \midrule
            \emph{Biomed \& CS} & SciCite & \citet{Cohan2019StructuralSF} & CLS & 85.49 & 84.76 $\pm$ 0.37 \\
            \midrule 
            \emph{Multi-domain} & PaperField & \citet{Beltagy2019SciBERTPC} & CLS & 65.71 & 65.99 $\pm$ 0.08 \\
            \bottomrule
        \end{tabular}
    }
    \caption{\gorbert test results are comparable with reported \scibert test results on the set of tasks and datasets from \citet{Beltagy2019SciBERTPC}, to which we refer the reader for descriptions. Reported statistics are span-level F1 for NER, token-level F1 for PICO, dependency parsing (DEP), and macro-F1 for relation (REL) and text (CLS) classification. We report micro-F1 for ChemProt. All \gorbert results are the mean $\pm$ standard deviation of 5 runs with different random seeds. \citet{Beltagy2019SciBERTPC} do not report standard deviation or number of runs.}
    \label{tab:replication_study}
\end{table*}

\section{Evaluation}
\label{sec:evaluation}

To evaluate the quality of our metadata selection, we randomly sample 500 paper clusters, restricting to those with PDFs. Within each sampled cluster, we determine whether the canonical title and authors match the title and authors in the selected canonical PDF.

Inline citation detection and bibliography parsing are dependent on \grobid \cite{Lopez2009GROBIDCA}. \citet{Ahmad2018CADAA} evaluate \grobid for detecting inline citations using a corpus of 5k CiteSeer papers, and found \grobid to have an F1-score of 0.89 on this task.  \citet{Tkaczyk2018MachineLV} report \grobid as the best among 10 out-of-the-box tools for parsing bibliographies, also achieving an F1 of 0.89 in an evaluation corpus of 9.5k papers. We perform an evaluation over 200 randomly sampled papers from \gorc and found comparable F1-scores for \grobid performance on both tasks.

For bibliography linking, we randomly sample \gorc papers (500 \grobid PDF parses and 100 \latex parses) and select one linked bibliography entry from each sampled paper (while avoiding selecting multiple entries linked to the same paper). We determine whether the title and authors in the bibliography entry agree with the title and authors of the linked paper. 

We present these evaluation results in Table \ref{tab:evaluation_results} and detail valuation criteria in Appendix \S\ref{appendix:evaluation}.

\begin{table}[ht!]
    \centering
    \begin{tabular}{lcc}
        \toprule
        Evaluated task & Title & Authors \\
        \midrule
        Paper clustering & 0.93 & 0.89 \\
        Bib. linking (\grobid) & 1.00 & 0.96 \\
        Bib. linking (\latex) & 1.00 & 0.92 \\
        \bottomrule
    \end{tabular}
    \caption{Accuracy of paper clustering and bibliography linking for titles and authors in sampled evaluation sets.}
    \label{tab:evaluation_results}
\end{table}

\section{Pretraining \bert on \gorc}
\label{sec:gorbert}

To demonstrate the suitability of \gorc for language model pretraining, we train \bert-Base \citep{Devlin2019BERTPO} on the parsed full text of \gorc and show that the resulting model (\gorbert) performs similarly to \scibert \citep{Beltagy2019SciBERTPC} on a diverse suite of scientific NLP tasks and datasets. 

While \scibert is a \bert-Base model also trained on multiple domains of scientific text, key differences in its pretraining corpus and vocabulary and those used for \gorbert are: 
\begin{itemize}[noitemsep]
    \item \textbf{Domain:} \citet{Beltagy2019SciBERTPC} report a pretraining corpus consisting of 82\% biomedical and 18\% computer science papers. Our \gorc pretraining corpus consists of a more balanced distribution of papers across diverse academic disciplines (see Figure~\ref{fig:fos_stats}), such that biomedical (42.7\%) and computer science (7.2\%) papers only comprise half the corpus.
    \item \textbf{Preprocessing:} \gorc identifies figure captions, table text and captions, headers, footers, and footnotes. We exclude these from the pretraining corpus. We tokenize and sentencize the text using scispaCy \citep{Neumann2019ScispaCyFA}. We also use heuristic filters to remove ill-formed paragraphs (such as those containing too many symbols).
    \item \textbf{Size:}  The resulting \gorc pretraining corpus contains 16.4B tokens, nearly five times larger than the corpus for \scibert.
    \item \textbf{Vocab:} Following \citet{Beltagy2019SciBERTPC}, we construct a cased WordPiece \cite{Wu2016GooglesNM} vocabulary of size 31k using 15\% of the \gorc pretraining corpus. The Jaccard index between the \gorbert and \scibert vocabularies is 0.536. 
\end{itemize}

We follow a similar setup to \citet{Beltagy2019SciBERTPC} for both pretraining and fine-tuning \gorbert.  Like \scibert, \gorbert is pretrained from scratch using the original \bert code\footnote{\url{https://github.com/google-research/bert}} and default \bert-Base configurations on a single TPU v3-8 for one week.  Also like \scibert, \gorbert is fine-tuned on all tasks by optimizing a cross entropy loss using Adam \citep{Kingma2014AdamAM}, a linear learning rate decay with 10\% warm-up, batch size of 32, and dropout of 0.1.  

We search over an equal-sized grid of hyperparameters as \citet{Beltagy2019SciBERTPC}. We fine-tune for 1 to 4 epochs with a maximum learning rate of 1e-5, 2e-5, 3e-5, or 5e-5. For each task, we select the optimal combination of these two hyperparameters using the development set and report the corresponding test set results.  For details, we refer the reader to \scibert code,\footnote{\url{https://github.com/allenai/scibert}} which we use for all experiments.

The results in Table~\ref{tab:replication_study} show that \gorbert outperforms \scibert on many tasks despite including a large percentage of data outside of the biomedical and computer science domains. 
As the pretraining corpus for \scibert is not publicly-available, \gorc can serve as a large pretraining corpus for evaluating and comparing pretraining approaches on academic text.  We also release \gorbert to serve as a baseline for research.

\section{Applications of \gorc}
\label{sec:discussion}

\gorc can be used for many NLP and analysis tasks over academic text. We give a summary of potential applications below.

The combination of structured full text annotated with linked inline citations makes \gorc well-suited for a variety of citation-related text-based tasks.
Without any additional supervision, \gorc can be used directly for both inline \citep{He2010ContextawareCR, Duma2014CitationRA, Jeong2019ACC} and document-level \citep{Yu2012CitationPI, Liu2015ContextBasedCF, Bhagavatula2018ContentBasedCR} citation recommendation.  Among document-level recommenders, \gorc is well-suited to the setting of \citet{Liu2015ContextBasedCF}, who use inline citation contexts to filter document-level recommendations. 

\begin{figure}[htb!]
    \centering
    \includegraphics[width=\columnwidth]{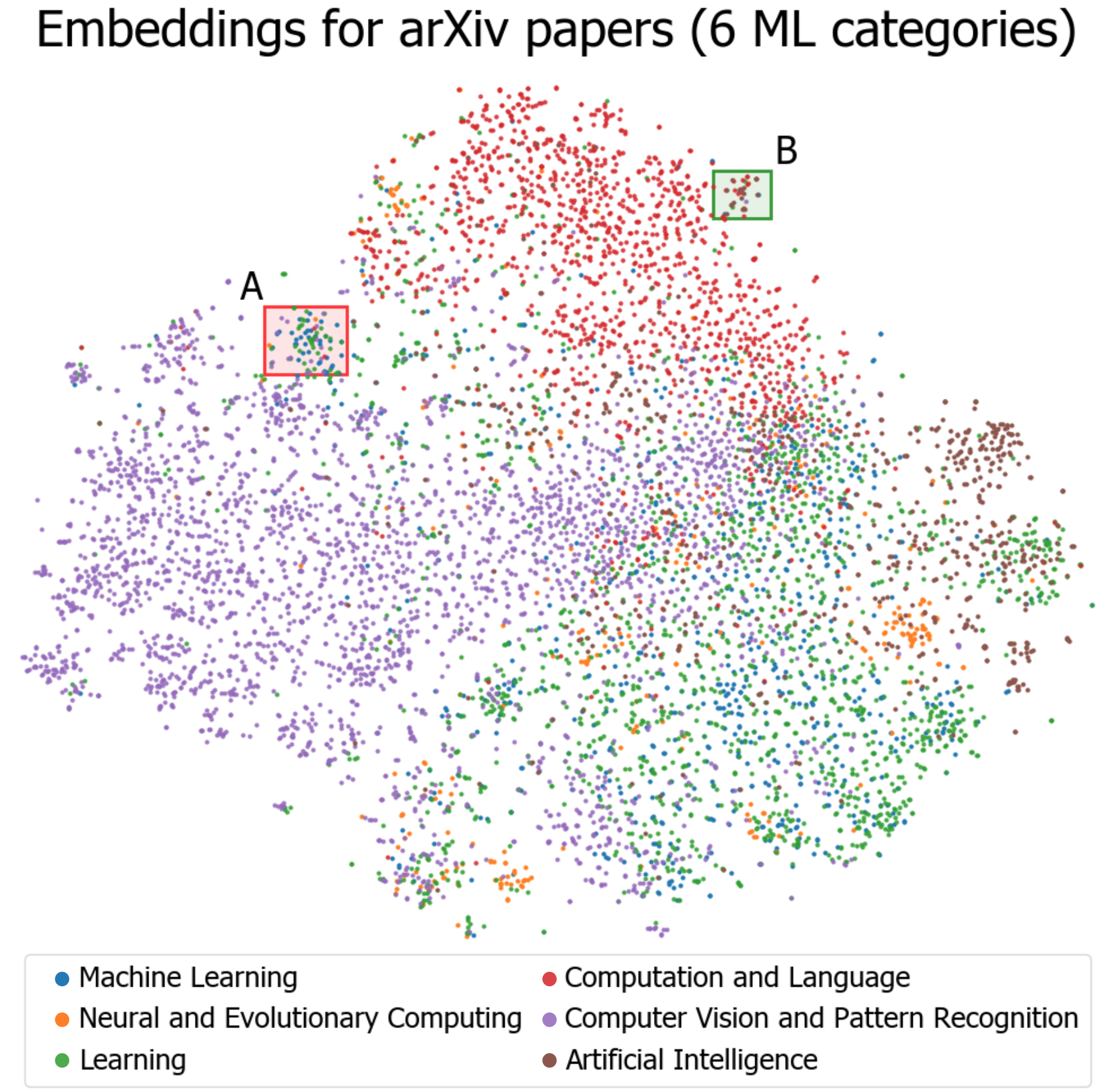}
    \caption{\emph{Word2vec} embeddings associated with 20k papers in six AI-related arXiv categories visualized using t-SNE \citep{Maaten2008VisualizingDU}. Example papers from two randomly selected sub-regions A and B are given in Table \ref{tab:tsne_groups}.}
    \label{fig:arxiv_cs_cats}
\end{figure}
\begin{table}[htb!]
    \centering
    \small
    \begin{tabular}{lp{55mm}}
        \toprule
        \multicolumn{2}{l}{Region A} \\
        \midrule
        cs.LG & ``On Unifying Deep Generative Models'' \\ 
        stat.ML & ``Learning Disentangled Representations with Semi-Supervised Deep Generative Models'' \\ 
        cs.LG & ``Denoising Criterion for Variational Auto-Encoding Framework'' \\ 
        cs.CV & ``Variational methods for conditional multimodal deep learning'' \\ 
        \midrule
        \multicolumn{2}{l}{Region B} \\
        \midrule
        cs.CL & ``TransA: An Adaptive Approach for Knowledge Graph Embedding'' \\ 
        cs.AI & ``TorusE: Knowledge Graph Embedding on a Lie Group'' \\ 
        cs.CV & ``Image-embodied Knowledge Representation Learning'' \\ 
        stat.ML & ``Neural Embeddings of Graphs in Hyperbolic Space'' \\ 
        \bottomrule
    \end{tabular}
    \caption{Sampled papers in clusters from t-SNE embedding space in Figure \ref{fig:arxiv_cs_cats}. Region A consists of papers related to deep generative models; region B consists of papers concerned with graph representation learning.}
    \label{tab:tsne_groups}
\end{table}

Other tasks that leverage citation contexts include classifying citation intent \cite{Teufel2006AutomaticCO, Jurgens2018MeasuringTE, Cohan2019StructuralSF}, identifying citation sentiment \cite{Athar2012ContextEnhancedCS}, identifying meaningful citations \cite{Valenzuela2015IdentifyingMC}, extracting key phrases \cite{Caragea2014CitationEnhancedKE}, and citation context-based paper summarization \cite{Teufel2006AutomaticCO, Qazvinian2008ScientificPS, Cohan2015ScientificAS, Mitrovic2015SummarizingCC}. The models in these papers require labeled citation contexts for training. \gorc could potentially benefit task performance without additional annotation, for example, by pretraining language models on \gorc citation contexts before fine-tuning to these tasks.  \citet{Cohan2019StructuralSF} find that long citation contexts (beyond sentence boundary) are important for tasks like summarization; the wider citation contexts available in \gorc could be used to augment existing datasets for document-level tasks.

Citation contexts can also be used for the more general tasks of identifying similar papers \cite{Kanakia2019ASH, Eto2019ExtendedCS, Haruna2018ACR, Small1973CocitationIT} or bibliometric analysis \cite{Ding2014ContentbasedCA, Trujillo2018DocumentCA, Asatani2018DetectingTI}. 
Towards these tasks, the citation contexts in \gorc can provide insight into \textit{how} and \textit{why} papers are cited. 
We illustrate this by following \citet{berger2016cite2vec} in training a \emph{word2vec} skip-gram model \citep{Mikolov2013DistributedRO} using full text citation contexts in \gorc, where each inline citation span is replaced with its linked paper identifier. When training over this modified text, the \emph{word2vec} model learns embeddings corresponding to each unique paper identifier, which can be leveraged as paper embeddings. The resulting embeddings shown in Figure \ref{fig:arxiv_cs_cats} and Table \ref{tab:tsne_groups} form clusters corresponding closely to arXiv Machine Learning categories.  Upon inspection, papers of different categories in the same embedding sub-region share research themes (see Table \ref{tab:tsne_groups}), indicating that these paper embeddings trained from citation contexts capture coherent topic similarity and relatedness. These paper embeddings can be used to identify similar papers, using the similarity between two papers' citing contexts as a proxy for paper similarity.

The \latex subset of \gorc also provides unique opportunities for research. In addition to citations and references, we also extract and parse tables from \latex source into a structured format. There is an opportunity to use these tables for corpus-level results extraction and aggregation. The \latex subset also has fine-grained extraction and labeling of mathematical formulas, which can be used to understand proof construction, or to assist in symbol co-reference resolution.

\section{Related work}
\label{sec:related_work}

The ACL Anthology Network (AAN) \cite{Radev:2009:AAN:1699750.1699759} is a bibliometric-enhanced corpus covering papers in the field of computational linguistics. It is built from the ACL Anthology \cite{bird-etal-2008-acl} and consists of 24.6k papers manually augmented with citation information. The PubMed Central Open Access corpus is a large corpus of 2.6M papers in the biomedical domain with citations linked to PubMed identifiers.\footnote{\url{https://www.ncbi.nlm.nih.gov/pmc/tools/openftlist/}} CiteSeerX \cite{Giles1998CiteSeerAA}, consists of papers collected primarily via web crawl, without integrating metadata provided by sources outside of the PDF. Although citation contexts are no longer available through CiteSeerX, the RefSeer dataset \cite{Huang2015ANP}\footnote{\url{https://psu.app.box.com/v/refseer}} is a dataset of short citation context snippets derived from 1.0M papers from CiteSeerX. More recently, \citet{Saier2019BibliometricEnhancedAA} introduce a corpus built using 1.0M arXiv publications. They use \latex source to extract text, citation spans and bibliography entries, which are linked to papers in the Microsoft Academic Graph. The citation context they provide are extracted snippets and no bibliography parses are provided. An updated version of this dataset \citep{Saier2020} released concurrently with this work now includes full text.

Compared with these resources, \gorc represents a significantly larger dataset of linked papers covering broad domains of science by leveraging PDF parsing in addition to \latex source. \gorc also provides clean full text for text mining and NLP needs with additional enhancements such as annotations of table and figure references and captions. 
\gorc's wealth of metadata and structured text allows it to be flexibly adapted to a variety of downstream tasks.

\section{Conclusion}

We introduce \gorc, the largest publicly-available corpus of English-language academic papers covering dozens of academic disciplines. \gorc consists of 81.1M papers, 380.5M resolved citation links, and structured full text from 8.1M open-access PDFs and 1.5M \latex source files. We aggregate metadata and abstracts from hundreds of trusted sources.  Full text is augmented with sections, citation mentions, and references to tables and figures. We demonstrate that \gorc can be used effectively for downstream NLP tasks in academic paper analysis. 

The pipeline for creating \gorc was used to construct the \cord corpus \cite{wang-lo-2020-cord19}, which saw fervent adoption as the canonical resource for COVID-19 text mining. \cord is aimed at assisting biomedical experts and policy makers process large amounts of COVID-19 literature in the search for effective treatments and management policies. With over 75K dataset downloads, dozens of search and question-answering systems, and hundreds of participating teams across two shared tasks\footnote{The Kaggle CORD-19 and TREC-COVID competitions. See \citet{wang-lo-2020-cord19} for details.} in the first month of its release, there is little doubt of the resource's impact. Our hope with the release of \gorc is to ensure such text mining resources are available to researchers even beyond periods of global crisis.

\section*{Acknowledgements}
This work was supported in part by ONR grant N00014-18-1-2193, and the University of Washington WRF/Cable Professorship.

We thank Doug Downey, Oren Etzioni, Andrew Head, and Bryan Newbold for their valuable feedback on the manuscript.  We also thank Isabel Cachola, Dallas Card, Mike D'Arcy, Suchin Gururangan, Daniel King, Rik Koncel-Kedziorski, Susan Liu, Kelvin Luu, Noah Smith, Gabi Stanovsky, and Dave Wadden for feedback on the dataset during early development.  Finally, we thank the Semantic Scholar team for assisting with data access and system infrastructure.

\bibliography{gorc_refs}
\bibliographystyle{acl_natbib}

\clearpage
\appendix

\section{Background \& Terminology}
\label{appendix:terminology}

In this work, we distinguish between bibliography entries and inline citations. A \textbf{bibliography entry} is an item in a paper's bibliography that refers to another paper. It is represented in a structured format that can be used for paper-identifying features such as title, authors, year, and venue or journal, and for journal articles, the volume, issue, and pages. Also commonly represented are unique document identifiers such as the Document Object Identifier (DOI), arXiv identifier, or PubMed identifier. Common formats for bibliography entries are MLA, APA, Vancouver-, and Chicago- style, among others, which are different ways of representing these various features for document identification.

There is often variation in the representation of certain fields. For example, \textit{Authors} can include the first names of each author or only their first initials. In many academic disciplines, journal publications are the norm, whereas conference proceedings dominate in fields such as Computer Science; conference proceedings tend to lack journal-related features such as \textit{Volume}, \textit{Issue}, and \textit{Pages}. Bibliography entry demarcation also varies between different formats. In some cases, each entry is preceded by a citation marker (e.g. ``[1]'' or ``[ABC2019]'') that is used throughout the text of the paper to denote inline citations. 

An \textbf{inline citation} is a mention span within the paper's abstract or body text that refers to one of the entries in its bibliography.

\begin{quote}
``\dashuline{ABC (2019)} present model 1, which outperforms model 2 \dashuline{(XYZ (2019))}.''
\end{quote}

In this example, the \textbf{narrative} inline citation \dashuline{ABC (2019)} appears as a noun phrase in the sentence while the \textbf{parenthetical} inline citation \dashuline{(XYZ, 2019)} is inserted into the sentence as an aside. A sentence remains grammatically correct when parenthetical citations are removed.  Other styles of parenthetical citations include, but are not limited to, \textsc{bracket}-style numbers (e.g. ``[1, 3-5]'') and \textsc{other} styles such as superscripts (e.g. ``$^{1, 2}$''), both of which refer to numbered entries in the bibliography. Bibliography entries without numbered entries or citation markers are typically referenced inline using \textsc{name-year} format as \dashuline{ABC (2019)} or \dashuline{(XYZ, 2019)} in the example above.

Additionally, an \textbf{inline reference} is a span in a paper that refers to another part of the paper, for example, references to figures, tables, equations, proofs, sections, or appendices. These often take on the form of:

\begin{quote}
    ``In \dashuline{Figure 3}, we show the relationship between A and B.''
\end{quote}

\noindent where \dashuline{Figure 3} refers to a plot displayed on a separate page. These inline references can be important for understanding the relationship between text and objects within the paper.

\section{PDF filters}
\label{appendix:pdf_filters}

Prior to running \grobid, we filter out PDFs that (i) produce an error when processed using the Python library PyPDF2,\footnote{Used to determine PDF page number and page dimensions} (ii) have greater than 50 pages (more likely to be a dissertation or report), (iii) have page widths greater than page heights (more likely to be slides), and (iv) those which fail to be extracted using pdfalto, the variant of pdftoxml used by \grobid. 

Numbers of PDFs removed by these filters are given in Table \ref{tab:pdf_filters}.

\begin{table}[ht]
    \centering
    \begin{tabular}{p{35mm}c}
        \toprule
        Filter & Number of PDFs \\
        \midrule
        PyPDF2 error & 0.54M \\
        Over 50 pages & 2.27M \\ 
        Page width $>$ height & 0.28M \\ 
        PDFAlto error & 0.21M \\
        \bottomrule
    \end{tabular}
    \caption{PDFs filtered out before \grobid processing}
    \label{tab:pdf_filters}
\end{table}

\section{The paper clustering problem}
\label{appendix:clustering_logic}

In academic fields in which preprint publishing is common (e.g. arXiv), the notion of a ``paper'' is somewhat ambiguous. For example, if a published paper differs from its arXiv preprint (as it often does), are the two documents considered separate papers for the purposes of citation? What about different arXiv preprint drafts tagged as different versions but under the same arXiv identifier?

In this work, each ``paper'' of interest is actually a collection (or cluster) of highly-similar (but not necessarily identical) documents. These paper clusters, provided by \semanticscholar, are constructed to reflect how authors tend to view their own papers; for example, most authors would consider their arXiv preprint and its associated published version to be the same ``paper''. For practical concerns in constructing \gorc, we further require that one document within the cluster be the canonical document used to represent the paper cluster.  

There are issues with defining a paper to be a collection of documents.  For example, suppose a paper cluster contains both an arXiv preprint and a peer-reviewed draft.  And suppose another paper cites the arXiv preprint critiquing content that has been updated in the peer-reviewed draft.  If the peer-reviewed draft is chosen as the canonical representation of the paper cluster, then the citation context would not accurately capture the rationale of that reference.  While worth noting, we believe such cases are rare and do not affect the vast majority of citation contexts.

\section{\gorc evaluation criteria}
\label{appendix:evaluation}
\paragraph{Paper cluster quality}  For each paper cluster, we compare the selected canonical \textit{Title} and \textit{Authors} fields with the title and authors of the selected canonical PDF. The \textit{Title} field is labeled correct if it exactly matches the title seen on the PDF, with some allowance for different capitalization and minor differences in special character representation (e.g. ``$\gamma$'' versus ``gamma'') and ignoring whitespace. The \textit{Authors} field is labeled correct if all authors on the PDF are presented in the correct order, with some allowance for variation in the surface form. This is to avoid penalizing publisher metadata for providing a first initial (instead of the first name) or omitting middle names or titles (e.g. ``Dr.'', ``PhD'').  

\paragraph{Paper-Bibliography linking} For each paper-bibliography pair, we compare the selected canonical \textit{Title} and \textit{Authors} fields in the structured bibliography entry to the selected canonical \textit{Title} and \textit{Authors} fields of the linked paper cluster.  The \textit{Title} fields are labeled as a match under the same criteria described above for matching paper cluster \textit{Title} fields and PDF titles.  The \textit{Authors} fields are labeled as a match if there is substantial overlap in the names of the authors.  For example, if authors A, B and C are in the bibliography entry and the linked paper cluster has authors A and B, then this is still considered a match. We note that in our evaluation, differences in the two sets of author names primarily stems from incorrectly written bibliography entries or mistakes in publisher-provided metadata.

\section{Training corpus sizes for other language models}

\begin{table}[tbh!]
    \centering
    \small
    \begin{tabular}{p{34mm}|p{33mm}}
        \toprule
        Language model & Training data \\
        \midrule
        \textsc{ELMo} \newline \cite{Peters2018DeepCW} & 1BW (800M) \newline Wikipedia (1.9B) \newline WMT 2008-2012 (3.6B) \\ [0.5mm]
        \hline \\ [-3mm]
        \bert \newline \cite{Devlin2019BERTPO} & BooksCorpus (800M) \newline Wikipedia (2.5B) \\ [0.5mm]
        \hline \\ [-3mm]
        \textsc{RoBERTa} \newline \cite{Liu2019RoBERTaAR} & BooksCorpus (800M) \newline CC-News (\textapprox 3.8B) \newline OpenWebText (\textapprox 1.9B) \newline Stories (\textapprox 1.6B) \\ [0.5mm]
        \hline \\ [-3mm]
        \textsc{GPT2} \newline \cite{Radford2019LanguageMA} & Web Text Corpus (\textapprox 2.8B) \\ 
        \bottomrule
    \end{tabular}
    \caption{Reported and estimated (several papers report corpus size in terms of bytes) token counts of training data used to train language models.}
    \label{tab:lm_training_data}
\end{table}

We estimate that all of \gorc consists of approximately 25B tokens of full body text and 15B tokens of abstract text. As demonstrated for \gorbert pretraining, aggressively-cleaned body text from the PDF-parsed subset of \gorc still yields approximately 16.5B tokens. The size of \gorc makes it more than sufficient for pretraining large language models such as \textsc{ELMo}, \bert, \textsc{RoBERTa}, \textsc{GPT2}, and others, whose reported training data sizes are given in Table \ref{tab:lm_training_data} for comparison.

\begin{figure}[thb!]
    \centering
    \includegraphics[width=\columnwidth]{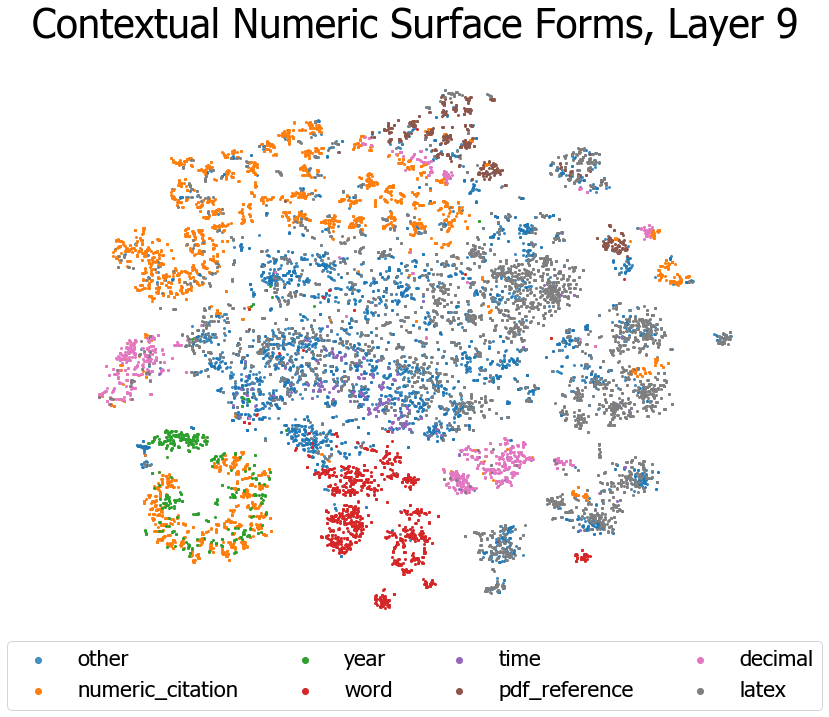}
    \caption{Visualization of contextual representations from layer 9 of \gorbert on numeric surface forms in a subsample of body text from \gorc. Labels are heuristics based on token-level patterns.}
    \label{fig:numeric_representations}
\end{figure}

\section{Numeric representations in \gorbert}
\label{sec:numeric_representations}

Academic papers contain substantially more diverse uses of numeric surface forms than typical web text, such as experimental results, equations, citation references and section/figure markers. To demonstrate this, we cluster contextual word representations involving numbers, heuristically labeling them into one of 8 categories based on surface patterns. Examining the progression of the contextual representations through the layers of \bert reveals an initial focus on sentence position (expected, due to explicit position embeddings) and magnitude, with later layers integrating substantial contextual information, such as the presence of inline \latex identifiers, citation indicators and PDF references. Following \citet{Peters2018DissectingCW, Liu2019LinguisticKA}, we observe that the final 2-3 \bert layers provide embeddings that excel at predictive language modeling; as such, Figure \ref{fig:numeric_representations} uses embeddings from layer 9 of \gorbert.

\section{MAG topic distribution}

\begin{table}[h!]
\begin{tabular}{lrr}
\toprule
\makecell[l]{\textbf{Academic} \\ \textbf{discipline}} & \makecell[l]{\textbf{All} \\ \textbf{papers}} & \makecell[l]{\textbf{Full text} \\ \textbf{available}} \\
\midrule
Medicine                     & 12.8M               & 1.8M                         \\
Biology                      & 9.6M                & 1.6M                         \\
Chemistry                    & 8.7M                & 484k                         \\
n/a                          & 7.7M                & 583k                         \\
Engineering                  & 6.3M                & 228k                         \\
Comp Sci                     & 6.0M                & 580k                         \\
Physics                      & 4.9M                & 838k                         \\
Mat Sci                      & 4.6M                & 213k                         \\
Math                         & 3.9M                & 669k                         \\
Psychology                   & 3.4M                & 316k                         \\
Economics                    & 2.3M                & 198k                         \\
Poli Sci                     & 1.8M                & 69k                          \\
Business                     & 1.8M                & 94k                          \\
Geology                      & 1.8M                & 115k                         \\
Sociology                    & 1.6M                & 93k                          \\
Geography                    & 1.4M                & 58k                          \\
Env Sci                      & 766k                & 52k                          \\
Art                          & 700k                & 16k                          \\
History                      & 690k                & 22k                          \\
Philosophy                   & 384k                & 15k        
                \\
\bottomrule
\end{tabular}
\end{table}

\end{document}